\documentclass[man,12pt]{apa7}
\usepackage{graphicx}

\usepackage{booktabs}
\usepackage{todonotes}
\usepackage[american]{babel}
\usepackage{csquotes}
\usepackage[style=apa,backend=biber]{biblatex}
\usepackage{censor}
\usepackage{tikz}
\usetikzlibrary{shapes.geometric} 
\usetikzlibrary{arrows,backgrounds,calc,hobby,positioning}
\ExplSyntaxOn
\cs_if_exist:NF \prg_stepwise_function:nnnN { \cs_gset_eq:NN \prg_stepwise_function:nnnN \int_step_function:nnnN }
\cs_if_exist:NF \prg_stepwise_inline:nnnn { \cs_gset_eq:NN \prg_stepwise_inline:nnnn \int_step_inline:nnnn }
\ExplSyntaxOff
\tikzset{vertex style/.style={
    draw=#1,
    thick,
    text=black,
    ellipse,
    minimum width=0.5cm,
    minimum height=0.25cm,
    font=\footnotesize,
    outer sep=1pt,
  },
  text style/.style={
    sloped,
    text=black,
    font=\footnotesize,
    above
  }
}

\usepackage{amsthm}
\newtheorem{thm}{Theorem}
\newtheorem{prob}[thm]{Problem}

\usepackage{hyperref}

\usepackage{subfigure}
\usepackage[nottoc]{tocbibind}
\usepackage{helvet}
\usepackage{courier} 
\usepackage{url}
\usepackage{prftree}

\usepackage{type1cm}         
\usepackage{amsmath}
\usepackage{makeidx}         
\usepackage{graphicx}        
\usepackage{multicol}        
\usepackage{tikz} 

\authorsnames{ Claudia Schon$^{1}$,, Ulrich Furbach$^{1}$, Marco Ragni$^{2}$}
\authorsaffiliations{{$^1$ Institute for Computer Science, University of Koblenz, 56070 Koblenz, Germany, \{uli,schon\}@uni-koblenz.de}, {$^2$ Predictive Analytics, Technical University Chemnitz, 09126 Chemnitz, Germany, ragni@informatik.uni-freiburg.de}}

\title{Modeling Associative Reasoning Processes}
\shorttitle{Modeling Associative Reasoning Processes}
\abstract{The human capability to reason about one domain by using knowledge of other domains has been researched for more than 50 years, but models that are formally sound and predict cognitive process are sparse. We propose a formally sound method that models associative reasoning by adapting logical reasoning mechanisms. In particular it is shown that the combination with large commensense knowledge within a single reasoning system demands for an efficient and powerful association technique. This approach is also used for modelling mind-wandering and the Remote Associates Test (RAT) for testing creativity. In a general discussion we show implications of the model for a broad variety of cognitive phenomena including consciousness.}
\begin{document}
\maketitle
%
%
%



\section{Introduction}

Associative thinking and reasoning, i.e., the capability to relate one concept with another by uncontrolled, sometimes subconscious thinking, is ubiquitous in our everyday life. It is a driving force for scientific progress, it is relevant for psychoanalysis, and cultural activities such as art are impossible to think without it \parencite{Scheffer2014}. Associative thinking is a core topic in cognitive psychology and has been systematically researched by the seminal work of Mednick for about half a century  \parencite[e.g.,][]{Mednick1962}, but only few and often domain restricted cognitive modeling approaches have been proposed so far \parencite{gawronski2014associative,benedek2012associative,DBLP:journals/kbs/OlteteanuSS19}. 

Consider a simple associative task taken from fRAT \parencite{fRAT}. 
\begin{prob}
Your task for the functional Remote Association Task (fRAT) is to find for three given words such as the following: 
\begin{center}
\emph{Tulip} ~~ \emph{Daisy} ~~ \emph{Vase} 
\end{center}
exactly one word that can be associated with them all in a meaningful way. For this example above an answer is \emph{Flower} (there can be other solutions too), because all three given words can relate to flower.
\end{prob}

In contrast to deductive or algebraic reasoning processes with a well-defined domain and operations,  problems such as the one above cannot be solved by applying just a set of known rules. Such problems require a form of insight, which is a ``sudden realization that appears to be correct'' \parencite{Webb2016} and creative thinking \parencite{DBLP:journals/kbs/OlteteanuSS19}. Creative processes can differ from individual to individual and there is no straightforward algorithm known that needs just be applied to solve such problems, and hence developing computational and cognitive models is difficult. General concepts need to be identified and compared and in principle there can be infinitely many of such concepts. As we can see from the problem above, semantic processes are relevant. Other  instances of such types of problems can be found in  commonsense reasoning. The follwing example is from a benchmark collection for commonsense problems:   

\begin{prob} \label{prob:copa}
\label{copa} An example taken from the Choice of Plausible Alternatives (COPA) Challenge \parencite{roemmele2011choice}:
\begin{center}

Assume that the pond froze over for the winter.

What happened as a result?

\begin{enumerate}
\setlength{\itemindent}{+.15in}
    \item People skated on the pond.
\item People brought boats to the pond.
\end{enumerate}
\end{center}
\end{prob}

 Another benchmark for commonsense reasoning with examples of this kind is the Winograd Schema Challenge \parencite{DBLP:conf/aaaiss/Levesque11}.  All these benchmarks have in common that it is not difficult for humans to come up with the correct response  together with an explanation. To tackle general problems such as the ones above, an extensive amout of background knowledge is necessary, from a cognitive modeling perspective. For Problem~\ref{prob:copa} this includes the knowledge that a frozen pond has a solid surface consisting of ice, that it is possible to skate on ice and that boats are usually not used on frozen ponds. Since commonsense reasoning problems cannot be assigned to a specific domain, it is not possible to represent all this knowledge manually which is one of the reasons why these problems currently pose real challenges for artificial intelligence (AI) (for more information, see \cite{Levesque2017}). Understanding the human ability to draw meaningful conclusions in everyday situations is therefore not only relevant for cognitive science, but for AI too. Understanding how we solve such everyday problems may tell us more about human cognition than well-defined abstract problems. The general task is to reverse-engineer \emph{how} associative reasoning can work on the cognitive level. We will focus here on Marr's computational level \parencite{Marr1982}. The reverse-engineering needs to be understandable -- in the sense of explainable AI. For this reason, we do not aim for a connectionistic or neural network based approach. We expect from a model for creative thinking to -- at least some extend -- whitebox the blackbox processes of associative thinking. This requires to use symbolic and knowledge rich modeling approaches. 

Since AI considers problems related to commonsense reasoning as challenging and largely unsolved, it is obvious, that there are no simple algorithmic realizations for such associative processes. Following the basic paradigm of `cognition is computation' and as long as we do not have computational models for general associative thinking, the gap between theories for associative thinking and cognitive models persists. To bridge this gap, in the following we will systematically introduce problems and domains such as commonsense reasoning and creativity thinking, among others and the current state of art in modeling. We demonstrate the importance of the need of a unified computational model for associative thinking that is (i) formally, as it is grounded in formal logics, (ii) able to approach concrete and core problems relevant for associative thinking, and (iii) providing new insights and hypothesis for the field. 

 Beyond answering the core research question -- \emph{Is it possible to model associative thinking by combining formally represented background knowledge, statistical methods, and automatic reasoning systems?} there are several more aspects this work will contribute too: It will provide a framework to analyze the impact of knowledge and inference processes and how different aspects are selected by cognitive processes. This framework is based on first-order predicate logic for the reasoning mechanism  and on various up-to date formalisms for very large knowledge bases. It will provide a testable, prototype implementation of the approach in Section~\emph{Associative Reasoning}. 
 It will apply the approach to 
 mind-wandering in Section~\emph{Mind-wandering} and in Section~\emph{Creativity} 
 human chains of associations as well as creative thinking are adressed. Furthermore the field of conscious and unconscious thinking will be touched in  Section~\emph{General Discussion}.

\section{Associative Thinking and Common Sense Reasoning}

Associative reasoning such as Problem 1 above is a part of creative thinking \parencite{Mednick1962,OLTETEANU201581}, as it requires ``to associate multiple remote items'' \parencite{OLTETEANU201581}. It  ``brings mutually remote ideas into contiguity 
facilitates creative solving” and believed that ``the organization of an individual’s associations will influence the probability and speed of attainment of a creative solution \parencite{Mednick1962}'' cited after  \parencite{OLTETEANU201581}. 
Connected to finding an associative word is \emph{abductive reasoning}, where the task is to find for observation(s) a best fitting explanation. Consider the following problem:

\begin{prob}
The following is from the abductive reasoning domain. Example from the aNLI (Abductive natural language inference) benchmarks \parencite{DBLP:conf/iclr/BhagavatulaBMSH20}.
Assume that two temporally successive observations are given:
\begin{itemize}
\setlength{\itemindent}{+.5in}
\item[$O_1$:] It was a very {hot summer day}.
\item[$O_2$:] He felt {much better}.
\end{itemize}
What is the more plausible hypothesis?
\begin{itemize}
\setlength{\itemindent}{+.5in}
    \item[$H^-$:] He decided to {run in the heat}.
\item[$H^+$:] He drank a glass of ice cold water.
\end{itemize} 
\end{prob}

For you as a human it is not difficult to imagine what could be the reason for $O_2$, but from a modeling perspective the underlying processes depend on knowledge and mental models. Formally, for an observation $O_1$ at time $t_1$ and an observation $O_2$ at time $t_2$ with $t_2 > t_1$, the task is to determine the more plausible explanation $H^+$ from the two given hypotheses $H^+$ and $H^-$. Abductive reasoning, the search for the best explanation and to  have  ``commonsense'' understanding about how the world may work are obviously connected.  Reasoning about plausibility is obviously connected to commonsense reasoning and requires some form of a model about the world. What fits this model of the world is \emph{plausible}. Another approach uses a core inference system, where the inference rules are obtained from transcriptions of various reasoning tasks that people have solved \parencite{Collins1989}. Their proposed system has two shortcommings: it has not been evaluated against a benchmark and the proposed system has been developed before knowledge became available in large quantities in a formalized manner and therefore does not address our core research problem -- how it is possible to find associated knowledge in large knowledge bases? Consider the following problem:
\begin{prob} Example from the SWAG dataset \parencite{DBLP:conf/emnlp/ZellersBSC18}.

On stage, a woman takes a seat at the piano. She
\begin{enumerate}
\setlength{\itemindent}{+.5in}
    \item [a)] sits on a bench as her sister plays with the doll.
    \item [b)] smiles with someone as the music plays.
    \item [c)] is in the crowd, watching the dancers.
    \item [d)] nervously sets her fingers on the keys.
\end{enumerate}
\end{prob}

This form of reasoning is clearly non-deductive, but these problems are very everyday-like. At the same time these presented  problems are part of cognitive tests and tasks spanning a variety of analogical and common sense reasoning both from psychology and AI \parencite{fRAT,DBLP:conf/emnlp/ZellersBSC18,DBLP:conf/iclr/BhagavatulaBMSH20,roemmele2011choice}. 

Another example are the problems in the Winograd Schema Challenge \parencite{DBLP:conf/aaaiss/Levesque11}. These problems consist of one sentence mentioning two parties and a reference to one of the parties with the help of a pronoun or possessive adjective. The task is to determine the party this pronoun or possesive adjective refers to. The difficulty results from the construction of these examples: The sentence always contains a special word where the replacement with an alternate causes the answer to switch. This is an example of a pair of problems in the Winograd Schema Challenge: 
\begin{prob} Examples from the Winograd Schema Challenge \parencite{DBLP:conf/aaaiss/Levesque11}:

The town councillors refused to give the angry demonstrators a permit because they feared violence. Who feared violence?
\begin{enumerate}
\setlength{\itemindent}{+.5in}
\item [Answer 0:] the town councillors
\item [Answer 1:] the angry demonstrators
\end{enumerate}

Using the alternate \emph{fear} for the special word \emph{advocated} leads to the following example:

The town councillors refused to give the angry demonstrators a permit because they advocated violence. Who advocated violence?
\begin{enumerate}
\setlength{\itemindent}{+.5in}
\item [Answer 0:] the town councillors
\item [Answer 1:] the angry demonstrators
\end{enumerate}
\end{prob}

Still other benchmarks aim to test a system's ability to continue telling a story. 
The Story Cloze Test is one of these benchmarks and is based on the ROCStories Corpora
\parencite{storyclozetest}. One ROCStory consists of a five-sentence story. An associated associated Story Cloze Test uses the first four sentences as a description of a situation and the task is to determine the next sentence from two alternatives, where the right choice is the original fifth sentence from the corresponding ROCStory. This is an example problem from the Story Cloze Test dataset:

\begin{prob} Example from the Story Cloze Test dataset \parencite{storyclozetest}. Given the four sentences below. Is A1 or A2 most likely to be the next sentence.

\begin{textit}
Karen was assigned a roommate her first year of
college. Her roommate asked her to go to a nearby
city for a concert. Karen agreed happily. The show
was absolutely exhilarating.
\begin{enumerate}
\setlength{\itemindent}{+.5in}
\item [A1:]  Karen became good friends with her roommate.
\item [A2:] Karen hated her roommate.
\end{enumerate}
\end{textit}
\label{cloze}
\end{prob}

Some of the presented problems above might be easier than others for a human reasoner in contrast to a formal system to deal with. What is common to all problems, however, is that without any background  knowledge, it is almost impossible to solve such problems. Problems that require knowledge are called \emph{knowledge rich problems}. But how do humans solve such problems on a cognitive level? 

This is still an open question. Despite a large number and variety of cognitive models for reasoning about knowledge lean problems such as for deductive reasoning, there are only few approaches for knowledge rich problems that associative models require. 

A way to integrate knowledge by using and adapting a Corpus of Contemporary American English (COCA) has been done in developing a cognitive model for the fRAT, the  comRAT-C \parencite{OLTETEANU201581}. Basically, for each expression in the problem, it searches the corpus to identify concepts. For this list of putative concepts, that can be connected to each other a selection is made. A variant of comRAT-c uses the frequency as an additional selector for the preferred response of participants. comRAT-c is able to solve 64 from the 144 problems. To the best of our knowledge this system has not been applied to the other types of problems presented above. 
Symbolical approaches steming from abductive and deontic reasoning, in other words approaches grounded in logic have been applied for solving a variant of COPA problems \parencite{Furbach2015}. In \textcite{10.1007/978-3-030-82147-0_11}, for example, the Winograd Schema Challenge problems are manually represented in the situation calculus, a variant of first-order logic. For each problem, matching knowledge is manually formalized and added. Then, again manually, it is decided whether abductive or deductive reasoning is more promising and then reasoning is performed.
In contrast, we aim at modeling the human ability to deal with large amounts of background knowledge. Therefore, we use pre-existing sources such as ontologies and knowledge graphs as background knowledge, from which we automatically extract knowledge suitable for a problem using associative selection.
In \textcite{triangle_copa}, the commonsense reasoning benchmarks \emph{Triangle Choice of Plausible Challenge (Triangle COPA)} are presented together with a solution approach based on weighted abduction. In these benchmark problems, situations, similar to the well-known experiments of \textcite{heider1944experimental}, are described and the task is to answer questions about these situations. The abduction-based approach to solve these problems also relies on manually formalizing suitable background knowledge for each problem and does not consider the human ability to deal with large amounts of background knowledge.

While recently  adverserial approaches \parencite{staliunaite2021improving} have been proposed, a core point is that models that can explain human plausible reasoning need some form of symbolic form to understand how these systems reason and use knowledge. 

\begin{table}[h]
\caption{An overview of different tasks and tests for associative thinking and common sense reasoning.}\label{overviewtests}
\begin{center}
    \begin{tabular}{lccc} \toprule
Task & Short Name &  Example  &  Citation\\ \midrule
Remote association test & fRAT     &1 & \parencite{fRAT}\\
Causal commonsense reasoning & COPA     & 2 & \parencite{roemmele2011choice}\\ 
Abductive natural language inference  & aNLI &  3 & \parencite{DBLP:conf/iclr/BhagavatulaBMSH20}\\
Grounded commonsense inference  & SWAG     & 4 & \parencite{DBLP:conf/emnlp/ZellersBSC18}\\
Winograd Schema Challenge  &      & 5 &  \parencite{DBLP:conf/aaaiss/Levesque11}\\
Story Cloze Tests  &      & 6 & \parencite{storyclozetest}\\
\bottomrule
\end{tabular}
\end{center}
\end{table}

Classical models for human reasoning dealing with large amounts of background knowledge do not represent this associative character and select background knowledge only based on syntactic equality of symbols. This completely ignores the meaning of symbols such as \emph{pond} in Problem 2 above and the context in which these symbols are used. However, when modeling human reasoning and associative thinking, it is precisely the meaning of symbols and the context in which they were used that is of great importance.
This is why, a combination of symbolic and statistical methods by integrating word similarities into the selection process is necessary. Word similarities allow us to model the associative nature of human thinking and gain context-specific background knowledge.

We can summarize that to solve the variety of problems of associative, commonsense, and abductive reasoning we need models that can (i) handle large amounts of background knowledge, (ii) select background knowledge in an associative way, (iii) comprises concepts and meaning, and (iv) is able to reason. In the following, we will present such a system.
Table~\ref{overviewtests} provides an overview of the benchmarks presented. For the development of our approach, we have focused on fRAT and COPA. This choice allows us to cover creativity with the fRAT problems and also to consider the area of causal commonsense reasoning, which is the topic of the problems in COPA.
However, the approaches we have developed can be applied to other benchmarks and we have also conducted preliminary experiments on the Story Cloze tests.

\section{Associative Reasoning}

\label{sec:reasoning}

Until now, we talked about various kinds of reasoning, like human, abductive or commonsense reasoning. 
Automated reasoning is one of the most traditional disciplines in Artificial Intelligence research. At the very beginning it was mainly aiming to prove mathematical theorems, but nowadays it is used as a major method in many other disciplines, like software and hardware verification, knowledge representation and even in modeling within cognitive science systems.
The latter is a topic tackling commonsense reasoning benchmarks like the COPA Challenge \parencite{roemmele2011choice} with the help of automated reasoning (\cite{DBLP:journals/ki/SchonSS19}). See Problem~\ref{copa} for an example problem from COPA. There are numerous other approaches to solve these problems with the help of statistical methods from machine learning, which offer good results \parencite{DBLP:conf/naacl/DevlinCLT19,RN+18,DBLP:journals/corr/abs-1904-09728}. 
As already mentioned in the introduction, the problem with these statistical methods is the lack of explanatory power. It is very difficult in these systems to get an explanation together with the given answer and therefore it is unclear what was actually learned \parencite{niven-kao-2019-probing}.  


As far as explainability is concerned, logic- or knowledge-based approaches are much better suited. The core within the system developed by  \textcite{DBLP:journals/ki/SchonSS19} is a classical first-order logic  theorem prover based on a tableau calculus. It will be discussed in this section for two reasons: Firstly, the prover is manipulating a single proof object, the tableau, which carries information about the state of the proof attempt at any time. In this way, it provides detailed information about the inferences performed at a given time. Secondly, the system is extended by a mechanism which allows to incorporate associative steps within very large knowledge bases which are connected to the system. The tableau and the control mechanism for these associations are explained in this section. It is important to note, that this first-order logical method can serve as a  core system for many variants of reasoning, like abductive  or deontic reasoning.  In the following subsection we depict this classical logical proof procedure and in another  subsection we show how to take the meaning of symbols into account for associative steps.

Later, in Section~\emph{General Discussion} 
we will discuss to which extend this entire system can be understood as a cognitive model for an information based understanding of consciousness.


\subsection{Reasoning: Logical Proof Procedure}
\label{sec:logicalreasoning}
The classical first-order logic system upon which the proposed commonsense reasoning system is build is Hyper, an automated theorem prover for first-order logic \parencite{BaumgartnerFurbachPelzer2007}. 
%

For this paper it is not necessary to understand the technical details of a reasoning system like Hyper, it should be enough to explain its functioning by means of an example.

We use the following example to illustrate the system.
\begin{prob}
\label{dog}
Example problem 65 from the COPA challenge: 
\begin{center}
The family took their dog to the veterinarian. 

What was the cause of this?
\begin{enumerate}
\item The dog chewed on a bone.
\item The dog injured his paw.
\end{enumerate}
\end{center}


\end{prob}

This natural language formulation of the problem is automatically translated into a logical formulation.

The right part of Figure~\ref{F:proof2} shows 13 formulae. Most of these formulae correspond to knowledge represented in ConceptNet \parencite{DBLP:conf/aaai/SpeerCH17}, some result from Problem~\ref{dog}, after the  translation  into first-order logic. 
The first 5 formulae are facts and the others are implications,  all of them are given in a normal form.

The left part of Figure~\ref{F:proof2} is a so called tableau, which is essentially a tree that was developed by Hyper with the help of inference rules. This tableau is a data structure which is extended by Hyper during its reasoning process.

 The technical aspects of the calculus on how to extend the tree in detail are not important here.  We want to point out, however, that
at any stage of the construction of a tree, a branch represents a (partial) interpretation of the given formulae, which means that it contains those facts which are assumed to be true. E.g., the right-most branch in our example corresponds to the (partial) interpretation 

$\{dog(a), bone(b),chew(c),on(c,b), \mathit{agent(c,a}), manducate(c), eat(c), $\\ 
$animal(a),carnivore(a), dog\_treat(b), \mathit{dog\_food}(b)\}$.

The left branch of the tableau is closed, since its nodes $bone(b)$ and $plant(b)$  are contradictory according to formula (11), which states that nothing can be a \emph{bone} and a \emph{plant}. 
A proof of the entire set of formulae is found if there is a tableau that contains only closed branches. If no proof can be found, like in the example in Figure~\ref{F:proof2}, the open branches list literals that can be derived from the set of clauses. The right branch contains both $\mathit{bone(b)}$ and $\mathit{dog\_food(b)}$ meaning that individual $b$ is a $\emph{bone}$ and also $\mathit{dog\_food}$.

Hyper has been used in many different application areas, reaching from commercial knowledge based systems 
to intelligent book development \parencite{DBLP:journals/jar/BaumgartnerFGS04}. 
Furthermore, Hyper was used as the main reasoning machinery in natural language query answering \parencite{DBLP:journals/aicom/FurbachGP10} and for cognitive reasoning, in particular answering commonsense questions \parencite{DBLP:conf/mates/FurbachS16}.

In all of these applications Hyper very rarely managed to find a proof within the given constraints --- in most cases there was a timeout and Hyper's result was a branch representing a partial interpretation of the formulae at hand. However, this is not a disadvantage, because in these cases Hyper can be used to draw inferences from a statement. Next, we will illustrate how to do this.
For the following  we use a running example from the aforementioned COPA benchmarks. As the example in Problem~\ref{dog} shows, these problems are formulated in natural language and require everyday knowledge to be understood and solved.

In order to draw inferences from a natural language statement with the help of an automated reasoning system, the problem has to be translated from natural language to a formal language --- in our case this is first-order logic. This translation is done in a fully automated system called KnEWS \parencite{knews}, which is based on  the Boxer-System  \parencite{CurranClarkBos2007ACL}.
As a running example, we consider the following sentence, which is a part of  Problem~\ref{dog}:
\begin{quote}
\emph{The dog chewed on a bone.}
\end{quote}
\noindent The first-order logic translation we get using KnEWS is:
\begin{align}
 \exists A (& dog(A) \land \exists B,C
                        (on(C,B)
                        \land bone(B)
                                             \land agent(C,A) 
                        \land chew(C)
                         )).\label{equ:chew}
\end{align}

While reading the above statement and the corresponding formula you certainly started thinking about a realistic scenario, e.g. about food intake of dogs or about the composition of meat and bones. We will discuss later how these scenarios may be invoked by association. Here, it is important to note, that we need knowledge in order to solve a task related to the statement.
Since our commonsense reasoning approach is intended to be able to handle problems from arbitrary domains, it is not possible to add background knowledge manually, as this would then have to be manually adjusted for each individual example. Therefore, our approach aims to model the human ability to deal with large amounts of background knowledge through associative reasoning. 
Therefore, we use existing  knowledge bases, where many possible facts and relations about the world are available. If this knowledge is not restricted to a single domain, if it is general enough to be used for different areas, it gets very large and hence difficult to handle. In \textcite{DBLP:journals/ki/SchonSS19}
among other sources ConceptNet \parencite{DBLP:conf/aaai/SpeerCH17} is used as background knowledge. ConceptNet is a semantic net structure and is actually multilingual. The English part contains  more than 1.5 million nodes.

Knowledge in ConceptNet is stored in the form of triples such as \emph{(dog, hasA, fur)}. Figure~\ref{fig:ex_conceptnet2} shows a small part of the information in ConceptNet connected to the words \emph{dog} and \emph{car}. To allow the first-order logic reasoner Hyper  to use ConceptNet as background knowledge, we have translated most of the English part of ConceptNet to first-order logic. The above triple has been translated into the following formula:
\begin{equation}
\forall X (dog(X) \rightarrow \exists Y (hasA(X,Y) \land \mathit{fur}(Y)))\label{equ:fur}
\end{equation}

The resulting knowledge base consists of 2,9 million axioms and is therefore far too large to be completely processed by reasoners.
Hence, it is necessary to select parts of this huge knowledge base which might be relevant for the task at hand. Note, that this situation is very different from a classical automated reasoning problem, 
where all the necessary formulae to find a proof are given and can be used all together  by the reasoning system, without the necessity to guess parts of it to be loaded into the system. 

The left part of Figure \ref{fig:selection} illustrates the situation in automated reasoning with large knowledge bases. The logical representation (Formula~(\ref{equ:chew})) of the natural language sentence is depicted on the very left together with the knowledge base, ConceptNet, in the middle of the figure. The task is to select those parts from the knowledge base which might be helpful for reasoning about the logical representation.

To this end there are two selection methods sketched: The first one uses syntactic criteria exclusively for the selection \parencite{Hoder:2011uq}. Depending on the symbols occurring within the logical representation those parts of the knowledge base are selected, which contain one of these symbols (additionally this selection takes the number of occurrences of a symbol into account in order to prevent that very frequent symbols like \emph{isA} lead to the selection of the whole knowledge base). Furthermore, depending on some parameters knowledge for other symbols occurring in the selected parts of the knowledge base can be selected.
The second selection method uses additional  semantic criteria for associative  selection; it is discussed in the following subsection.

\subsection{Associative Selection: Meaning of Symbols}

\label{sec:meaning}
\begin{figure*}
\begin{center}
\includegraphics[scale=0.45]{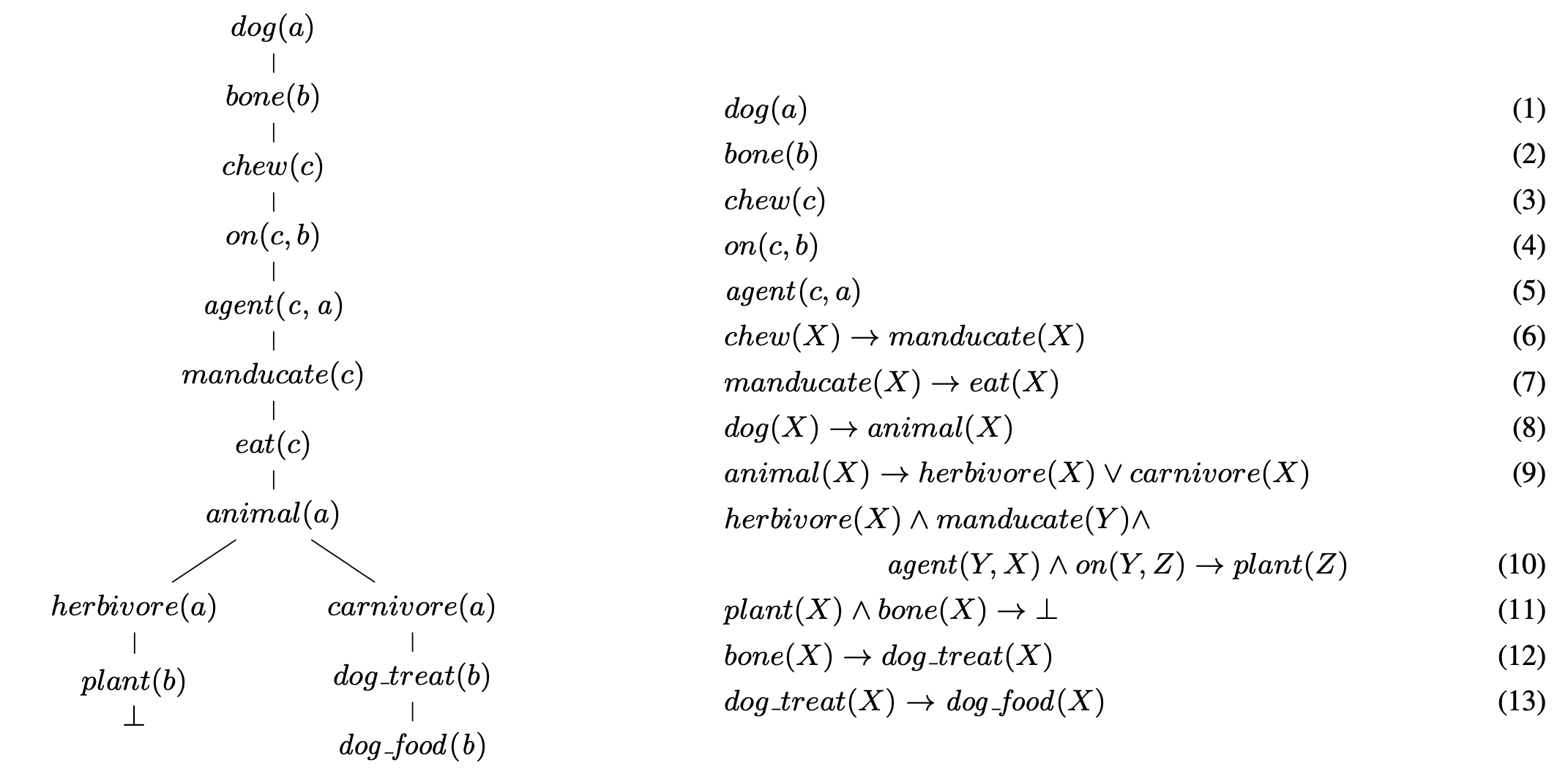}
\end{center}
\caption[bla]{\small Formulae on the right, hyper tableau for the formulae on the left. 
The left branch of the tableau is closed, the right branch is open. The literals in the open branch constitute a partial interpretation of the formula set.
    }
\label{F:proof2}
\vspace{-0.35cm}
\end{figure*}

In logical reasoning the symbols and names within a formula have no a priori meaning. In our example from Figure~\ref{F:proof2} the predicate name \emph{dog} could be changed to something completely different, say \emph{cat} or \emph{p}, the logical meaning of the entire set of formulae would be exactly the same, provided the change would be done consistently within all formulae. And indeed this is a very important property of logical reasoning: the logical structure of a formula is of interest --- all reasoning is done independent of the used signature (i.e.\, symbol names). The situation is completely different if a human reads the formula (provided she is not a logician). It is very likely that the symbols  \emph{dog}, \emph{chew} and \emph{bone}  trigger a certain scenario in the reader --- she reads the formula with the triggered interpretation\footnote{whereas in logical reasoning properties of a formulae are proven with respect to all possible interpretations}. There are numerous results from cognitive science, which indicate that such an assignment  of meaning to symbols is important for human reasoning. The Wason selection task \parencite{wason1968reasoning}, as an example, is an experiment where test persons have to solve a simple problem which basically is a logical implication. Most persons fail if the question is formulated in an abstract logical way, whereas if they  receive the same task, but formulated with much more everyday context, they perform much better. Another example is the semantic priming effect: the processing of a stimulus can be influenced by a preceding stimulus. The word \emph{dog} for example can be more efficiently processed, if the preceding word was \emph{animal} in contrast to something semantically unrelated, like \emph{planet}.

We use such a semantic similarity as a base for our associative steps aiming at a  selection of  formulae from the knowledge base, that may be helpful for the reasoning process.
 The semantics of ``a word is characterized by the company it keeps'' \parencite{Firth} --- which is the basic idea of distributed semantics of language, as it was popularized by Firth already in the beginning of the 20th century.
Recently it became obvious that this kind of semantics is very useful for statistic based machine learning.  Word embeddings 
offer an implementation for such semantics. Based on very large corpora of natural language text the words are  related to other words which occur in a certain neighborhood. This results in a multidimensional vector space which relates words that are used together very frequently close together --- this leads to a kind of similarity measure for words. 
In our model, different word embeddings come to use: The ConceptNet Numberbatch \parencite{DBLP:conf/aaai/SpeerCH17} word embedding, pre-trained GloVe word vectors\footnote{\label{note1}GloVe word vectors Common Crawl (840B tokens, 2.2M vocab, cased, 300d vectors, 2.03 GB download at \url{https://nlp.stanford.edu/projects/glove/})} and  a word embedding  learnt on personal stories from blog entries \parencite{roemmele2011choice}. ConceptNet Numberbatch is a word embedding combining and enhancing several pretrained embeddings with knowledge  from ConceptNet.


 The semantics of  symbols within our formulae is given by a word embedding, which we can use to find semantically similar symbols for the selection process. As a result not only formulae containing the symbols \emph{dog}, \emph{chew} and \emph{bone} are selected, but also those containing similar symbols like for example \emph{manducate} and \emph{remasticate}.
This method is described and evaluated in detail in 
\textcite{DBLP:conf/cade/FurbachKS19}. 

\begin{figure*}
\begin{center}
\includegraphics[scale=0.35]{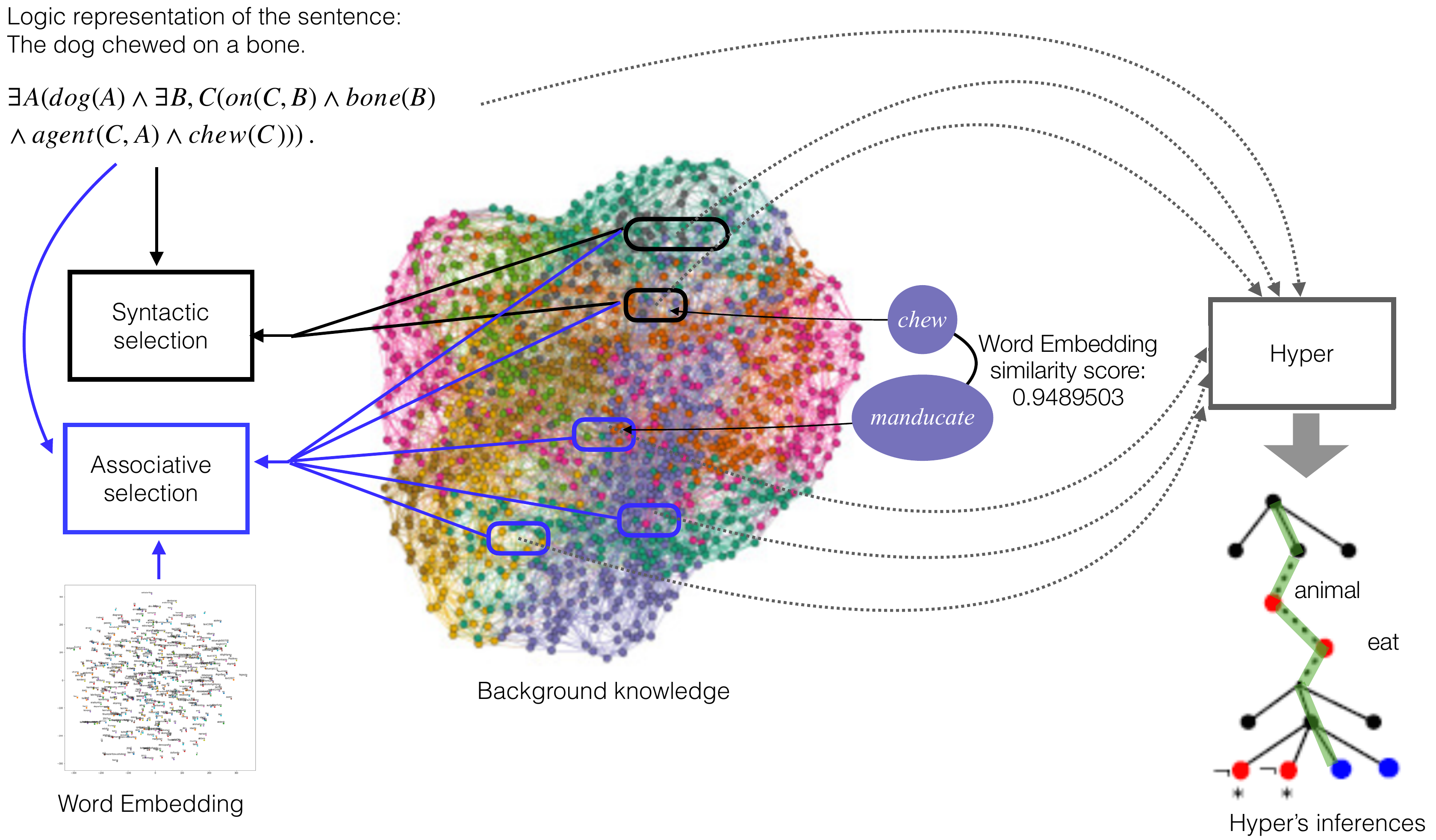}
\end{center}
\caption{On the left: Syntactic selection uses symbols from the formula to select parts of the background knowlege, depicted with black arrows and regions. Associative selection takes the meaning of symbol names  into account by additionally selecting formulae containing symbols which are similar according to a word embedding (depicted by blue arrows and regions). On the right: A snapshot during a Hyper run. The (green) path of the tree, Hyper is working on, can be interpreted as  the working memory; whereas the long-time
 memory is representated by the knowledge base. \\
 (Information on source of some parts of the graphics: Picture of network: \parencite{DBLP:journals/asc/GibsonV16}, CC BY 4.0 (\url{https://creativecommons.org/licenses/by/4.0/}).
Visualization of Word Embedding: \emph{Euskara: Hitz batzuen errepresentazioa} by Aelu013, CC BY-SA 4.0 (\url{https://creativecommons.org/licenses/by-sa/4.0/deed.en}) (word removed).)}
\label{fig:selection}
\vspace{-0.35cm}
\end{figure*}

Figure~\ref{F:proof2}
depicts an extract of the selected background knowledge for Formula~(\ref{equ:chew}) on the right hand side:  Formulae (6) - (13) correspond to selected background knowledge for the symbols in Formula~(\ref{equ:chew}) and exemplary for the symbol \emph{manducate} which is similar to \emph{chew}. Formulae (1) - (5) correspond to the normal form  of Formula~(\ref{equ:chew}).

Even if much background knowledge is added, this background knowledge will never be able to represent the complete human background knowledge and will always remain incomplete. Therefore, automatic theorem provers can only rarely prove statements or entailements  in natural language. For example, it would hardly be possible with an automatic theorem prover to prove that the statement \emph{The dog chewed on a bone} implies the statement \emph{The dog is content}. This is because the second statement is not  a logical consequence of the first one. It is rather the case, that \emph{The dog is content} is more likely to be a  consequence than the statement \emph{The dog is injured}.
This kind of reasoning is also called cognitive reasoning.



 Therefore, we do not aim at the construction of proofs but rather to analyse the inferences performed by Hyper  after a certain amount of reasoning (see the green path in the right part of Figure~\ref{fig:selection}) together with those parts of the knowledge which is selected by associations. 
 
As we have seen, one of the main problems in the above depicted commonsense reasoning task is the selection of appropriate parts of knowledge. 
Within the knowledge base we  have two different kinds of networks or connection layers. One layer is given by the structure of the knowledge base. E.g. two concepts \emph{dog} and \emph{mammal} may be connected by a relation \emph{IsA}, or \emph{car} and \emph{engine} by the relation \emph{HasA}, as it is depicted in Figure~\ref{fig:ex_conceptnet2}.
\begin{figure} 
\begin{center}
\begin{tikzpicture}[node distance=2cm]
	 \node[vertex style=black] (bone) {bone};
   	 \node[vertex style=white, left=0.5cm of bone] (q1) { }
 	 edge [-,cyan!60!blue] node[text style]{$\vdots$} (bone);
  	 \node[vertex style=white, above left=0.5cm of bone] (q2) { }
 	 edge [-,cyan!60!blue] node[text style]{} (bone);
  	 \node[vertex style=white, below left=0.5cm of bone] (q3) { }
 	 edge [-,cyan!60!blue] node[text style]{$\vdots$} (bone);
	 \node[vertex style=black, right=2.0cm of bone] (fur) {fur};
  	 \node[vertex style=white, right=0.5cm of fur] (p1) { }
 	 edge [-,cyan!60!blue] node[text style]{$\vdots$} (fur);
  	 \node[vertex style=white, below right=0.5cm of fur] (p2) { }
 	 edge [-,cyan!60!blue] node[text style]{$\vdots$} (fur);
  	 \node[vertex style=white, above right=0.5cm of fur] (p3) { }
 	 edge [-,cyan!60!blue] node[text style]{} (fur);
	\node[vertex style=black, below=2.0cm of bone] (Skf) {dog}
	 edge [->,cyan!60!blue] node[text style]{HasA} (fur)
 	 edge [->,cyan!60!blue] node[text style]{Desires} (bone);
   	 \node[vertex style=black, left=1cm of Skf] (d1) {mammal}
 	 edge [<-,cyan!60!blue] node[text style]{isA} (Skf);
  	 \node[vertex style=white, below left=1cm of Skf] (d2) { }
 	 edge [-,cyan!60!blue] node[text style]{$\vdots$} (Skf);
  	 \node[vertex style=white, below=1cm of Skf] (d3) { }
 	 edge [-,cyan!60!blue] node[text style]{$\vdots$} (Skf);
	\node[vertex style=black, below=2.0cm of fur] (Cf) {poodle}
	 edge [<-,cyan!60!blue] node[text style]{RelatedTo} (Skf);
	   	 \node[vertex style=white, right=0.5cm of Cf] (po1) { }
 	 edge [-,cyan!60!blue] node[text style]{$\vdots$} (Cf);
  	 \node[vertex style=white, below right=0.5cm of Cf] (po2) { }
 	 edge [-,cyan!60!blue] node[text style]{$\vdots$} (Cf);
  	 \node[vertex style=white, above right=0.5cm of Cf] (po3) { }
 	 edge [-,cyan!60!blue] node[text style]{} (Cf);
  	\node[vertex style=black, right=0.5cm of po1] (car) {car};
    \node[vertex style=black, right=2.0cm of car] (engine) {engine}
    edge [<-,cyan!60!blue] node[text style]{HasA} (car);
    \node[vertex style=white, right=0.5cm of engine] (engine_1) { }
 	 edge [-,cyan!60!blue] node[text style]{$\vdots$} (engine);
  	 \node[vertex style=white, below right=0.5cm of engine] (engine_2) { }
 	 edge [-,cyan!60!blue] node[text style]{$\vdots$} (engine);
  	 \node[vertex style=white, above right=0.5cm of engine] (engine_3) { }
 	 edge [-,cyan!60!blue] node[text style]{} (engine);
    \node[vertex style=black, above=2.0cm of car] (garage) {garage}
    edge [<-,cyan!60!blue] node[text style]{AtLocation} (car);
    \node[vertex style=white, left=0.5cm of garage] (garage_1) { }
 	 edge [-,cyan!60!blue] node[text style]{$\vdots$} (garage);
  	 \node[vertex style=white, below right=0.5cm of garage] (garage_2) { }
 	 edge [-,cyan!60!blue] node[text style]{$\vdots$} (garage);
  	 \node[vertex style=white, above right=0.5cm of garage] (garage_3) { }
 	 edge [-,cyan!60!blue] node[text style]{} (garage);
    \node[vertex style=black, right=1.5cm of garage] (convertible) {convertible}
    edge [<-,cyan!60!blue] node[text style]{IsA} (car);
    \node[vertex style=white, right=0.5cm of convertible] (convertible_1) { }
 	 edge [-,cyan!60!blue] node[text style]{$\vdots$} (convertible);
  	 \node[vertex style=white, below right=0.5cm of convertible] (convertible_2) { }
 	 edge [-,cyan!60!blue] node[text style]{$\vdots$} (convertible);
  	 \node[vertex style=white, above right=0.5cm of convertible] (convertible_3) { }
 	 edge [-,cyan!60!blue] node[text style]{} (convertible);
    \node[vertex style=white, left=0.5cm of car] (c1) {}
    edge [-,cyan!60!blue] node[text style]{$\vdots$} (car);
    \node[vertex style=white, above left=0.5cm of car] (c2) {}
    edge [-,cyan!60!blue] node[text style]{} (car);
    \node[vertex style=white, below left=0.5cm of car] (c3) {}
    edge [-,cyan!60!blue] node[text style]{$\vdots$} (car);
	\end{tikzpicture}
	\caption{Small part of the information in ConceptNet for the words \emph{dog} and \emph{car}. In reality, node \emph{dog} e.g.  has more than $660$ ingoing and $610$ outgoing edges. }\label{fig:ex_conceptnet2}
\end{center}
\end{figure}

In Figure~\ref{fig:selection} this layer is the dense and colorful network. The second layer is defined by the word embedding, which relates concepts according to their occurrences within a text corpus, e.g. \emph{dog} may be related to \emph{betting} if the text corpus on which the word embedding was learnt is about dog races. This second layer corresponds to associative connections between words, it is depicted by the purple bubbles which show the associations.
Paying attention to these associative links allows us to model the associative nature of human processing of large amounts of background knowledge in our selection.
All in all, we have a highly complex network consisting of these different connections with the logical reasoning  system Hyper using the different connection layers. 
 \color{black}
In the following sections, this entire system is applied to various associative tasks. 

\section{Mind-wandering}
\label{sec:mindwandering}

Mind-wandering in humans takes place when a person is awake, turned away from the outside world, thinking of nothing concrete, indulging in daydreams. Associative thinking plays an important role in mind-wandering were thoughts typically move from one concept to another in an uncontrolled manner. Neuroscientists have shown that in these states free from external stimuli, the so-called idle or default mode network is active \parencite{Otti:2012ly}. These mental time-outs provide an opportunity to spontaneously run through thoughts, produce new perspectives, simulate scenarios that can be helpful for processing the past or planning actions. When the person's attention is taken up again by a task, the energy of the default mode network is lowered so that he or she can concentrate on the task at hand. 
A study  \parencite{killingsworth:gilbert:2010} argues that up to 40\% of the time a human mind is wandering around.  

Mind-wandering also has interesting positive effects such as finding creative solutions to a problem, which has been shown in \textcite{article}. In this section we show that a variant of the approach introduced in this paper  can be used to model the process of mind-wandering. We denote this variant as \emph{mind-wandering system}. 


\subsection{Overview of the Mind-wandering System}
The mind-wandering process is started from an initial formula, such as Formula~(\ref{equ:chew}) from Section~\emph{Associative Reasoning}. In the first step, the system performs a associative selection as described in the previous section to select suitable background knowledge for this formula. The formula together with the selected background knowledge is transferred to Hyper, which performs inferences and returns a (possibly partial) interpretation. This  interpretation corresponds to the green path of the tree in the right part of Figure~\ref{fig:selection} and contains everything Hyper was able to derive within a given time limit. Since the selected background knowledge is very broad, the interpretation also contains very broad information. To find a focus, the symbols occurring in the interpretation are clustered according to their word meaning and a cluster is selected as focus. 
For the symbols in the focus, new background knowledge is selected and passed to Hyper again together with the focus. Hyper again performs inferences. This process can be repeated as many times as desired.
A detailed description of the individual steps of the system follows.

\subsection{Background Knowledge and Associative Selection}
 The  associative selection from the knowledge base  starts with a set of symbols called the current \emph{context}, which consists of the symbols from a starting formula like for example Formula~(\ref{equ:chew}) and similar symbols. Adding the similar symbols from the associative selection steps ensures that the current \emph{context} also represents the associative links in the knowledge base. We use this context as a running example in this section.

Selecting all formulae from the knowledge base in which one of the context symbols occurs results in a large set of formulae. Using all these formulae would be too unfocused w.r.t. the considered context, so a filtering step removes all formulae in which other symbols occurring in the formula are not within a certain range of similarity to the symbols in the context. To measure similarity, cosine similarity in a word embedding is used. 
The interval in which the similarity must fall for a formula to be selected is passed to the system by two parameters. With the help of these parameters it is possible to control how far the background knowledge is allowed to move away from the context symbols.  With a suitable interval it is possible to select  a formula like (\ref{equ:fur}) which represents knowledge about dogs and fur while preventing to select a formula like
\begin{equation}
\forall x (poodle(x) \rightarrow \exists y (\mathit{relatedTo(x,y)} \land dog(y)))\label{equ:poodle}
\end{equation}
Currently, the system can use either the ConceptNet Numberbatch \parencite{DBLP:conf/aaai/SpeerCH17} word embedding or a word embedding  learnt on personal stories from blog entries provided by \textcite{roemmele2011choice}.

\subsection{Reasoning}
The selected set of formulae together with  Formula~(\ref{equ:chew}) is passed to the Hyper reasoner. Hyper is started with a timeout of 30 seconds and calculates during this time a possibly partial interpretation for the input formulae. This interpretation represents knowledge that can be inferred from Hyper's input. In the next step, the system analyses Hyper's output. Since the input formulae are still very broad despite the filter methods mentioned above, the (partial) interpretation also contains very broad knowledge inferred from Hyper's input. 
First the mind-wandering system extracts all predicate symbols from Hyper's output and removes from this set all symbols from the current context to prevent the mind-wandering process from getting stuck. Hyper's model produced for the running example contains 122 symbols which are thematically widely spread: from \emph{ears, skin, flesh, wolf, calcium, animal, collar} and \emph{vertebrate} to \emph{woof} and \emph{barking}, many terms are represented.

\subsection{Finding a Focus}
To determine a focus in the multitude of these symbols, the mind-wandering system performs a clustering on these symbols using KMeans and the cosine similarity of a word embedding as similarity measure. Currently, the number of clusters created corresponds to the number of  symbols in the (partial) interpretation divided by 4. In future work, different values for the number of clusters will be considered.
Next, the system orders the resulting clusters by their cosine similarity to the  symbols in the current context and chooses one of the clusters as the focus. 
\begin{figure*} 
\begin{center}
\includegraphics[scale=0.5]{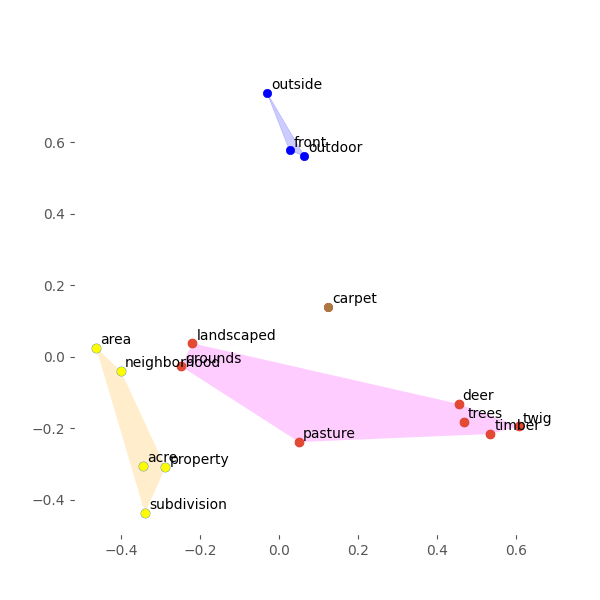}
\end{center}
\caption[caption]{Symbols occurring in a Hyper model in a step of a mind-wandering chain. The different polygons correspond to the clusters produced by k-means clustering. The word \emph{carpet} is a cluster with only one element. The figure was generated by automatically reducing the 300-dimensional ConceptNet Numberbatch word vectors to 2d vectors.}
\label{fig:clustering}
\vspace{-0.35cm}
\end{figure*}

The choice of the cluster to work with next is related to associative thinking. The thematic distance between the chosen cluster and the current context of the mind-wandering chain corresponds to the difference between concepts that are related to each other during associative thinking.
For the experiments, the cluster in the middle of the sorted sequence is chosen as the focus in order to allow the mind-wandering process to move away from the current context.
Other choices are possible and can be used to simulate different variants of associative thinking (with very strong or rather weaker connections between the concepts related to another).
In the running example, in the first iteration of the mind-wandering process this led to selecting the cluster consisting of the symbols \emph{animal} and \emph{animals} as the new focus. Figure~\ref{fig:clustering} shows the symbols occurring in one of Hyper's models corresponding to a later step in the mind-wandering process. The different colored polygons represent the clusters formed for these symbols. In the example, four clusters were created (one for each colored polygon and one for the one-worded cluster \emph{carpet}). The cluster consisting of the symbols \emph{outside}, \emph{front} and \emph{outdoor} was selected in this case. Other choices for the focus cluster are possible and can be selected with the help of a parameter.
After selecting a cluster, the system creates a simple formula from the symbols in the focus cluster and performs associative selection of background knowledge suitable for this formula as described above. In this selection, the symbols from the focus cluster together with similar symbols are used as new context symbols.
The described process 
is repeated a desired number of times or until the process does not deliver any new symbols.

\subsection{Experimental Results}Starting from the symbols in the initial formula, the symbols in the selected focus clusters represent the result of the mind-wandering process.
Starting from Formula~(\ref{equ:chew}) containing the symbols \emph{dog}, \emph{chew} and \emph{bone} the described system provides for example the following sequence of sets of focus symbols:
\begin{align}
1:\{&\mathit{dog, chew, bone}\}\nonumber\\
2: \{&\mathit{animal, animals}\}\nonumber\\
3:\{&\mathit{gardening}\}\nonumber\\
4:\{&\mathit{garden, horticulture, farming}\} \nonumber\\
5:\{&\mathit{mowing, lawn, yard}\} \nonumber\\
6:\{&\mathit{outside, front, outdoor}\} \nonumber\\
7:\{&\mathit{weather}\}\nonumber\\
8:\{&\mathit{thunder, lightning}\} \nonumber\\
9:\{&\mathit{cloud, sky, clouds}\} \nonumber\\
10:\{&\mathit{water}\}\hphantom{covering, pattern, textile, fastened, worn, wool, coat, material}\nonumber
\end{align}
This corresponds to a mind-wandering chain which focuses on animals  leads  to gardening and finally addresses weather aspects which leads to water.

It should be noted that the system has many parameters to control this mind-wandering process. For the experiments, different parameter combinations were automatically tried out and the sequences of focus symbols generated in this way were manually inspected. 
Different parameter values led to a different chain which finally ends at fashion:
\begin{figure*} 
\begin{center}
\includegraphics[scale=0.4]{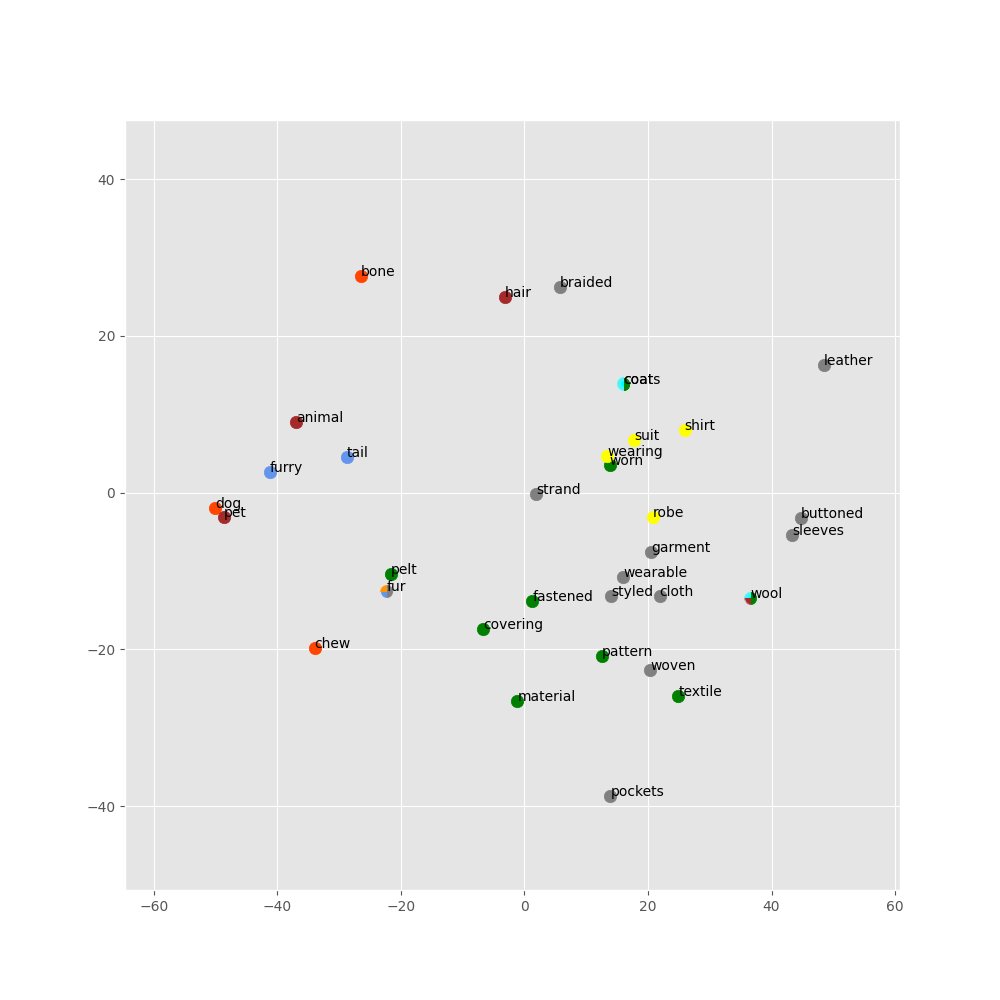}
\end{center}
\caption[caption]{Illustration of a mind-wandering chain. Each color represents the symbols of one step of the mind-wandering chain. The nodes representing \emph{fur}, \emph{coats} and \emph{wool} have multiple colours because they belong to more than one step of the mind-wandering chain. For example \emph{wool} belongs to steps 3, 5 and 7. The figure was generated by automatically reducing the 300-dimensional ConceptNet Numberbatch word vectors to 2d vectors.}
\label{fig:chain_fashion}
\end{figure*}
\begin{align}
1:\{&\mathit{dog, chew, bone}\} \nonumber\\
2:\{&\mathit{furry, tail, fur}\} \nonumber\\
3:\{&\mathit{coats, wool, coat}\} \nonumber\\
4:\{&\mathit{fur}\} \nonumber\\
5: \{&\mathit{animal, pet, hair, coat, pelt, wool}\} \nonumber\\
6:\{&\mathit{sleeves, robe, braided, leather, fur, garment, buttoned, styled, pockets, } \nonumber \\
& \mathit{strand, woven, cloth, wearable}\} \nonumber\\
7:\{&\mathit{coats, covering, pattern, textile, fastened, worn, wool, coat, material, pelt}\}\nonumber\\
8:\{&\mathit{wearing, robe, suit, shirt}\}\nonumber
\end{align}
Figure~\ref{fig:chain_fashion} illustrates this mind-wandering chain. Each color represents one step of the mind-wandering chain with all its elements. 

The presented experiments are only a first feasibility study. 
In addition to that, we did first experiments of the application of mind-wandering in the commonsense reasoning area where we consider commonsense reasoning benchmarks like the Choice of Plausible Alternatives Challenge \parencite{roemmele2011choice}.  
To solve these benchmarks we start a mind-wandering process for both answer candidates and to choose the answer where the result of the mind-wandering process is closer to the sentence \emph{The family took their dog to the veterinarian}. The distance between a mind-wandering chain and a sentence could be measured for example using cosine similarity of the correspondig vectors of a word embedding.
One example for a mind-wandering chain produced by the mind-wandering process is: 
\begin{align}
1:\{&\mathit{dog, injured, paw}\} \nonumber\\
2:\{&\mathit{hurt}\} \nonumber\\
3:\{&\mathit{injured, ouch, bruise, harm, painful, injure, ache}\}\nonumber\\
4:\{&\mathit{twinges, discomfort, dangerous, contusion, hurt, wound}\}\nonumber\\
5:\{&\mathit{incision, stitches, gauze, scar, healing, scab, bandaged, cuts}\}\nonumber\\
6:\{&\mathit{regenerate, tissue, scalpel, crusted, skin, bleed, scratch, fungus, wound}\}\nonumber\\
7:\{&\mathit{disease}\}\nonumber\\
8:\{&\mathit{infection}\nonumber\}
\end{align}
Compared to the two mind-wandering chains starting from \emph{The dog chewed on a bone.} presented above, this mind-wandering chain is clearly closer to the sentence \emph{The family took their dog to the veterinarian}. However, this is only one of the mind-wandering chains produced by the mind-wandering process. There are other mind-wandering chains that do not point in the direction of a visit to the veterinarian like the following one:
\begin{align}
1:\{&\mathit{dog, injured, paw}\}\nonumber\\
2:\{&\mathit{hurt}\}\nonumber\\
3:\{&\mathit{limb, injure, dislocated, wounded}\}\nonumber\\
4:\{&\mathit{disfigure, damage, harm, damaged, maimed, hurt, wound}\}\nonumber\\
5:\{&\mathit{battle, fight}\}\nonumber\\
6:\{&\mathit{boxing, contest, struggle, conflict, crusade, oppose, action}\}\nonumber\\
7:\{&\mathit{military}\}\hphantom{, action, action, action, actionaction, action, action, action}\nonumber
\end{align}

Therefore, in future work we plan to always generate several mind-wandering chains for each answer alternative and then select the answer whose mind-wandering chains are on average closer to the desired sentence.

\section{Creativity}

\label{sec:creativity}


This section applies our technique of associative selection in large knowledge bases to modelling creativity tasks. A very common test  to measure human creativity is the Alternate Uses Task \parencite{AUT}: Subjects are asked to find as many alternative uses as possible for certain objects such as a brick in a given time window. It is obvious that for the solution of such a task knowledge about the object plays a crucial role: the fact that a brick is heavy, hard, stable and cuboid-shaped represents important background information. Moreover, associative thinking is used to link this information with the properties of other objects in order to find an alternative use for the brick.

Another possibility to measure creativity constitute the Remote Associates Test (RAT)  \parencite{RAT1,RAT} and functionally Remote Associates Test (fRAT) \parencite{fRAT}.
In RAT, three words like \emph{cottage}, \emph{swiss} and \emph{cake} are given and the task is to find a fourth word that connects these three words. For this example, the solution is \emph{cheese}, since there is \emph{cottage cheese}, \emph{swiss cheese} and \emph{cheese cake}. 
fRAT is similar to RAT such that it also presents three words and the task is to find a fourth connecting word. In contrast to RAT, however, a functional relationship between the three words and the target word has to be found. For example, for the words  \emph{tulip},  \emph{daisy},  \emph{vase} the connecting word \emph{flower} has to be found.
To solve the problems in RAT and fRAT, a broad knowledge is needed. The solution of the above fRAT task requires the knowledge that \emph{tulips} and \emph{daisies} are flowers. In addition, the subject must have the knowledge that a \emph{vase} is a container in which flowers are kept. The difficulty in these tasks is to find a connection in the knowledge belonging to the three given words. Humans rely on associative thinking to find this connection.

Therefore, in order to model human creativity in AI systems, these systems must be able to combine knowledge from different domains and have to be able to model associative thinking. 
In everyday life, the human ability to connect knowledge even goes so far that people easily come up with an explanation for seemingly unrelated observations: 
\begin{quote}
Imagine you see some snow and a carrot next to a sleigh. 
\end{quote}
\noindent The obvious explanation, that the snow and the carrot are the remains of a melted snowman, requires the combination of knowledge from different fields. For an AI system to come up with such an explanation, a purely syntactic selection of background knowledge is not sufficient in many cases, because the knowledge selected in this way is not broad enough. Syntactic selection would probably end up selecting the information that carrots are vegetables, snow is a type of precipitation and a sleigh is a vehicle. However the connection between these three words would probably not be selected. This example illustrates that a component modelling associative thinking is mandatory to enable AI systems to come up with comparable explanations.

Even if we furthermore select knowledge for words similar to snow, carrot and sleigh, we might end up missing the connection to a snowman. Table~\ref{tab:sim} lists for each of the words \textit{snow}, \textit{carrot} and \textit{sleigh} the most similar words found in the pre-trained GloVe word vectors\footnote[6]{\label{note2}GloVe word vectors Common Crawl (840B tokens, 2.2M vocab, cased, 300d vectors, 2.03 GB download at \url{https://nlp.stanford.edu/projects/glove/})}.
\begin{table} 
\begin{tabular}{ll} 
\toprule
\textbf{word}\phantom{dd} & \textbf{10 most similar words}\\ 
\midrule 
\textit{snow} & \textit{winter, snowfall, snowy, snows, rain, snowing, Snow, snowstorm,}\\
			 & \textit{weather, sleet}\\
\textit{carrot} & \textit{carrots, celery, potato, onion, zucchini, cabbage, broccoli, cauliflower,}\\
		& \textit{turnip, cucumber}\\
\textit{sleigh} & \textit{Sleigh, sleighs, sled, reindeer, horse-drawn, reindeers, trundle, seldge,}\\
 & \textit{sleds, headboard}\\
\bottomrule
\end{tabular}
\caption{10 most similar words for \emph{snow}, \emph{carrot} and \emph{sleigh} according to pre-trained GloVe word vectors. Words are sorted in descending order according to their similarity to the respective word in the left column. }\label{tab:sim}
\end{table}
Even if we select knowledge for all the similar words listed in Table~\ref{tab:sim} the knowledge about snowmen will probably not be selected.

Therefore, it is important that the available information (the three words \emph{snow}, \emph{carrot} and \emph{sleigh}) is first linked to a context and knowledge is selected for this context.
However, determining this context is not trivial. One commonly used way would be to calculate the mean value for the word vectors of the words of interest (\emph{snow}, \emph{carrot} and \emph{sleigh}) and then determine the most similar words for this mean value (like done in \cite{abels2021focusing}).
This results in the following 20 words (sorted in descending order according to their distance to the mean value):
\begin{align}
&\mathit{pumpkin, snowman, carrots, reindeer, winter, snowy,  } \nonumber \\
&\mathit{sledding, sled, potato, ice,snowflakes, potatoes, turnip, }\nonumber \\
& \mathit{cabbage, apples, autumn, wintry, frosty, bunny, pumpkins}\label{equ:snowman}
\end{align}
\begin{figure*} 
\begin{center}
\includegraphics[scale=0.4]{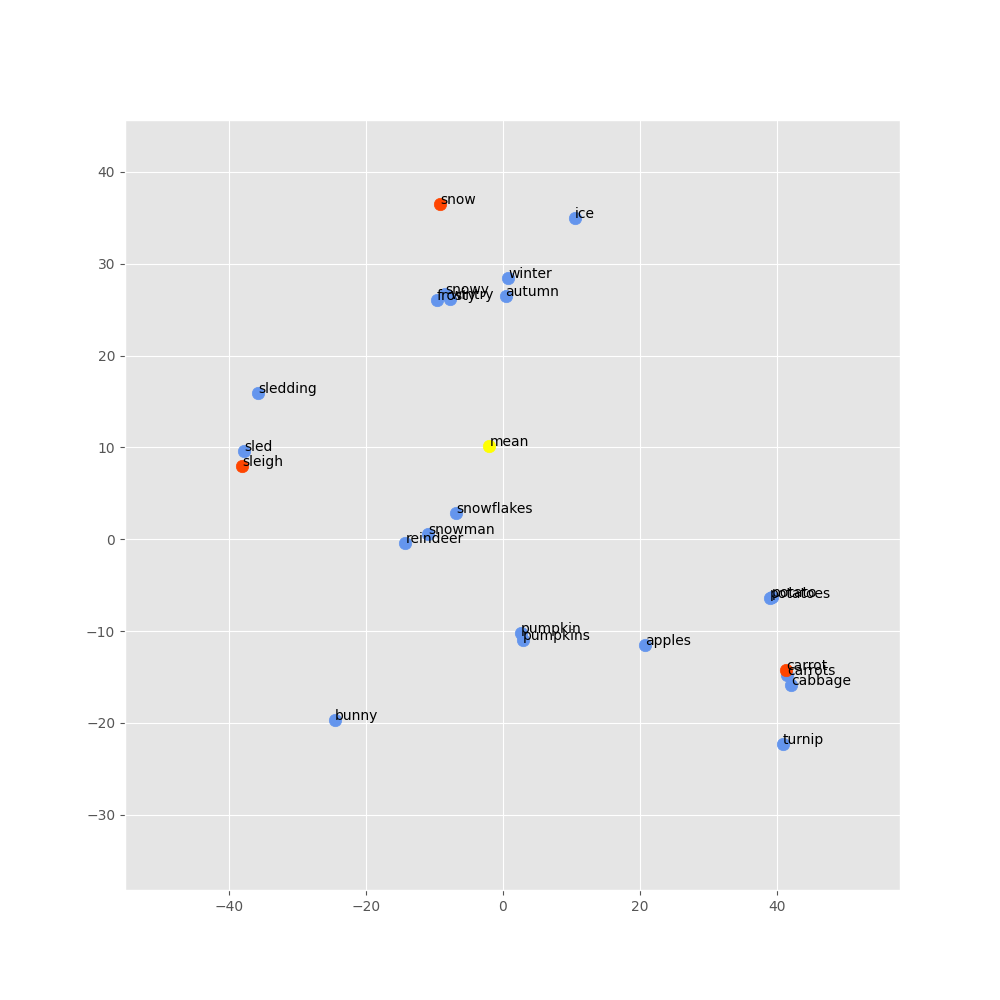}
\end{center}
\caption[caption]{Illustration\footnotemark[7] of the words most similar to the mean of the context words \emph{carrot}, \emph{sleigh} and \emph{snow}. The three context words are depicted in red, the mean in yellow. The words depicted in blue correspond to the 20 words most similar to the mean and are listed in Equation~\ref{equ:snowman}.}
\label{fig:snowman}
\end{figure*}
As we can see, the snowman already appears in second place in this set.
 Even though this is an approach to determining context that is often chosen, it is immediately noticeable that the result contains many unwanted words that are obviously not part of the context. Among the 20 words listed, 8 are vegetables or fruits. If we determine the distances of the vegetables and fruits to the three context words (\emph{snow}, \emph{carrot}, \emph{sleigh}) we notice that they are only close to \emph{carrot} and the distance to \emph{snow} and \emph{sleigh} is quite high. Figure~\ref{fig:snowman} illustrates nicely, that the  vegetables and fruits have a high closeness to the mean determined from the context words, but they have a high variance as far as the closeness to the individual context words is concerned.
Since it is actually desirable that the words determined are close to the mean and have low variance with respect to proximity to each context word, we include variance in experiments to determine which words belong to a context.

In experiments we determined the 100 words which are most similar to our context words and sorted them by the difference of the distance to the mean of the context words and the variance to the individual context words. For the running example with the context words \emph{snow}, \emph{carrot} and \emph{sleigh}, this results in the following words (sorted in descending order according to the difference):
\begin{quote}
\textit{snowman, pumpkin, reindeer, sledding, winter, ice, snowy, snowflakes, sled, bunny, pumpkins, pine, snowflake, autumn, frosty, marshmallow, christmas, snowmen, snowball, carrots}
\end{quote}

These words only contain three vegetables. The \emph{snowman} appears at the first place of the word list. Furthermore, new words interesting for the context like \emph{snowball}, \emph{snowflake} and \emph{christmas} appear in the word list. Compared to the context words presented above, this second list of words provides a better context for the words \emph{snow}, \emph{carrot} and \emph{sleigh}.

In first experiments, we tested the usefulness of word embeddings for solving the f-RAT benchmarks. As mentioned above, each f-RAT problem consists of three query words (like \emph{discuss}, \emph{gossip}, \emph{telephone}) together with the answer (in our example \emph{talk}). For each of the 48 f-RAT problems considered in \textcite{DBLP:journals/kbs/OlteteanuSS19}, we considered the query words and determined the 200 words most similar to the mean of the query words in the GloVe word vectors. We sorted these 200 words in descending order of similarity to the mean of the three query words. 
For 28 of the 48 fRAT problems, we were able to find the answer word we were looking for among the first 10 similar words. 
Impressively, the answer word could even be found among the first 3 similar words in 23 of 48 cases.
In addition, we observed that sorting the similar words according to the difference suggested above between distance to the mean of the query words and the variance of the similarity to the individual query words resulted in the answer word appearing higher up in the sorting in 21 of the 48 problems. Only in 7 of the 48 problems did this sorting result in the answer word appearing further down in the sorting.

Table~\ref{tab:exp} shows the experimental results for the first five f-RAT problems given in \cite{DBLP:journals/kbs/OlteteanuSS19}. We list the 5 words most similar to the mean of the query word as well as the 5 words which are most similar to the difference of the mean und the variance. The answers given in \textcite{DBLP:journals/kbs/OlteteanuSS19} are given in bold face. It is noticeable that not only the correct answer but also other similar words listed in the table make sense as an answer for the query. For example for the query words \emph{sensitive, sob} and \emph{weep}, the correct solution is \emph{cry}. The word \emph{tear}, which is second among the candidates for answers we have determined, also represents a possible solution.

But even in cases where the answer word was not found among the first 10  similar words, we found words which could be a solution as well. For instance, for the query words \textit{arrest, badge} and \textit{deputy} the answer \textit{cop} was not found among the top 10 words, but the first two hits \textit{officer} and \textit{police} could be a solution as well.
\begin{table} 
\begin{tabular}{ll} 
\toprule
\textbf{query words}\phantom{dd} & \textbf{most similar words: (a) most similar to mean, }\\
&\textbf{(b) most similar to difference of mean and variance}\\ 
\midrule 
\textit{question, reply, solution} & \textit{(a) \textbf{answer}, questions, suggestion, answers,problem}\\
			 & \textit{(b) \textbf{answer}, suggestion, answers, problem, questions}\\
\textit{sensitive, sob, weep} & \textit{(a) \textbf{cry}, tears, sigh, moan, groan}\\
		& \textit{(b) \textbf{cry}, tears, sigh, crying, moan}\\
\textit{antlers, doe, fawn} & \textit{(a) antlers, \textbf{deer}, elk, moose, fawn}\\
 & \textit{(b) \textbf{deer}, elk, antlers, moose, whitetail}\\
 \textit{bud, dandelion, petals} & \textit{(a) petal, blossom, leaf, blossoms, \textbf{flower}}\\
 & \textit{(b) blossom, petals, leaf, blossoms, \textbf{flower} }\\
  \textit{colt, mare, unicorn} & \textit{(a) stallion, filly, foal, gelding, \textbf{horse}}\\
 & \textit{(b) stallion, foal, filly, \textbf{horse}, gelding}\\
\bottomrule
\end{tabular}
\caption{Experimental results for the first 5 f-RAT problems: Query words together with the 5 words most similar to the mean of the query words. Furthermore, the 5 words highest difference of similarity to mean of the query words minus the variance of the distances to the mean. Similarities determined using pre-trained GloVe word vectors. The gold standard answer is displayed in bold.}\label{tab:exp}
\end{table}


Altogether these  first experiments are encouraging and show that the inclusion of values other than the distance to the mean vector could be useful. In future work, we plan to investigate this in more detail.

\section{General Discussion}
\label{sec:general}
\label{sec:consciousness}
In the previous sections we have discussed mind-wandering and creativity from the perspective of associative thinking and we discussed our associative reasoning system as a cognitive model for these phenomena. But mind-wandering and creativity also seem to be closely related to consciousness.  \textcite{baars2010spontaneous} argues that  mind-wandering  is closely related to conscious life-relevant problem solving and learning. The link between creativity and consciousness seems very obvious. In the literature, there is even a technique called ``stream of consciousness'' that attempts to mimic the character's thought process (James Joyce used this technique prominently in Ulysses). In this section we broaden the discussion by adressing  aspects of consciousness in relation to the associative reasoning system used until now. 

Consciousness is a topic in several scientific disciplines, such as philosophy, psychology, biology or neuroscience. In this paper, we cannot cover all aspects of consciousness research, but rather focus on one branch of research that has an obvious connection to our system: Consciousness from an information-theoretical point of view. For this discussion we use Chalmers' seminal paper \parencite{chalmers1995facing} as a starting point. This paper classifies problems around the notion of consciousness into easy and hard problems. Easy problems include explanations of certain phenomena, e.g. descriptions of mental states or the focus of attention or the ability of a system to access its own internal state. According to Chalmer the really hard problem is that of \textit{experience} or \textit{qualia}. This notion can be circumscribed by \textit{something it is like to be}, as it is formulated prominently by \textcite{10.2307/2183914}. Chalmers is proposing several research methods aiming at tackling this hard problem of consciousness. In articular he is proposing an information theory approach which  was later taken up by Tononi:



\subsection{Information Integration}

\begin{figure*} 
\begin{center}
\includegraphics[scale=0.4]{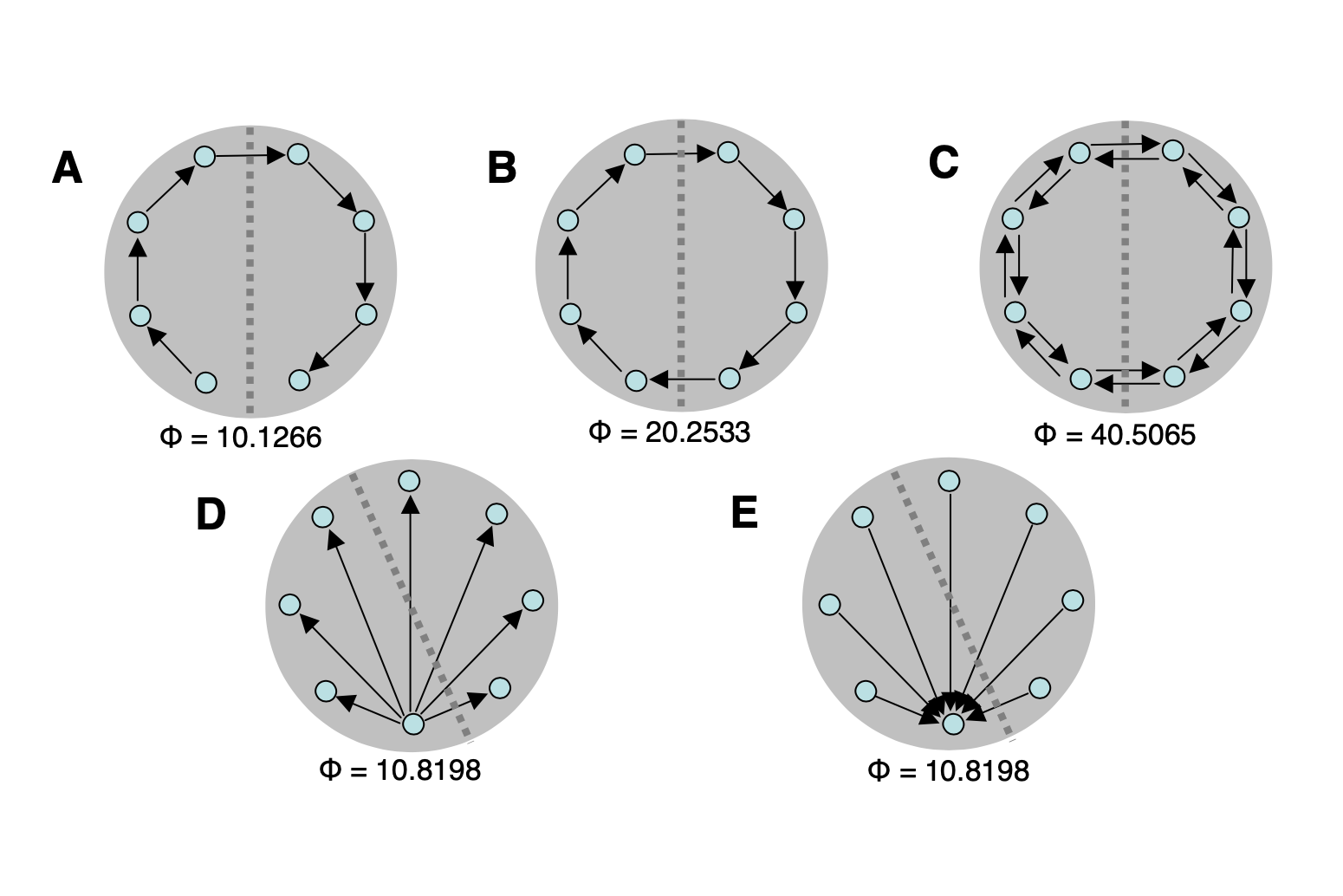}
\end{center}
\caption{Information integration for basic digraphs. (A) Directed path. (B) One-way cycle. (C) Two-way cycle. (D) Fan-out digraph. (E) Fan-in digraph. Complexes are shaded and values for $\Phi$ are provided in each of the panels.(Image and caption is taken from \parencite{Tononi:2003se})}
\label{fig:tononi-graphs}
\end{figure*}

Integrated Information  Theory of \textcite{tononi:04} avoids the necessity of a neurobiological correlate of consciousness. It is applicable to arbitrary networks of information processing units, they need not be neural or biological. Tononi is proposing a thought experiment: Assume you are facing a blank screen, which is alternatively on and off and you are instructed to say ``light'' or ``dark'' according to the screen's status. Of course a simple photodiode can do exactly the same job, beep when the light is on and silence when it is off. The difference between you and the photodiode is the so called ``qualia''  --- you consciously experience ``seeing''  light or dark. This is a partially  subjective process, a first-person feeling, which we are not able to measure or compare  with that of other persons (a prominent treatment of this topic is in \cite{10.2307/2183914}). One difference between you and the photo diode is that the diode can switch between two different states, on and off, exclusively whereas your brain enters one of an extremely large number of states when it recorgnizes the light. But it is not just the difference in the number of states, it is also important to take the degree of information integration into account. 
If we use a  megapixel camera instead of a single photodiode for differentiating light from dark, this technical device would also enter one of a very large number of possible states (representing all possible images it can store). According to Tononi the  difference between you and the camera, is that the millions of pixels within the camera are not connected to each other. Your brain, however, handles a huge number of information and it integrates  information from various parts of the brain. (For example it is hard to imagine colours without shapes.) \textcite{tononi:04} offers a formal definition of information integration by defining a function $\Phi$, which measures the capacity of a system to integrate information. 
To get an idea of this approach, assume a network of elements which are connected, e.g. a neural network and take a subset $S$ from this system. Tononi ``wants to measure the information generated when $S$ enters a particular state out of its repertoire, but only to the extent that such information can be integrated, i.e. it can result from causal interactions within the system. To do so, we partition $S$ into $A$ and its complement $B$~\ldots We then give maximum entropy to the outputs of $A$, i.e. substitute its elements with independent noise sources of constrained maximum variance. Finally, we determine the entropy of the resulting responses of $B$ \ldots''  \parencite{tononi:04}. Based on this computation  he  defines the \emph{effective information between $A$ and $B$}, which is a measure  of the information shared between the source $A$ and the target $B$. The above mentioned function $\Phi$ is defined with the help of the notion of effective information. The entire system is then divided into several bipartitions in order to find the ones  with the   highest  $\Phi$-value. These so-called complexes are responsible for integration of information within the system. 
Tononi is considering them  as the ``subject'' of experience, being the locus where information can be integrated.
It should be clear, that the computation of complexes for realistic networks is highly complex and not really feasible. For some basic graph structures the $\Phi$-values are given in Figure~\ref{fig:tononi-graphs}.

Based on this understanding of consciousness, one can try to test the theory by considering  several neuroanatomical or neurophysiological factors that are known to influence consciousness of humans. Tononi does this in great detail in \textcite{tononi:04}, others are developing new Turing tests for AI systems based on this theory (e.g. \cite{Koch:2008}).
It is not surprising that Integrated Information Theory is criticized from various directions.
With respect to the notion of information Searle \parencite{Searle:2013}   is arguing that information according to Shannon is observer-dependent, it is `only' a theory about syntax and encoding of contents, whereas consciousness is ontologically subjective and observer-independent. Koch and Tononi's reply  \parencite{Koch:Tononi:2013}  is that they use ``a non-Shannonian notion of information—integrated information —- which can be measured as ``differences that make a difference'' to a system from its intrinsic perspective, not relative to an observer. Such a novel notion of information is necessary for quantifying and characterizing consciousness as it is generated by brains and perhaps, one day, by machines.'' It is  important to note, that
Integrated Information Theory offers a way to quantify the degree of consciousness of a system.
This is exactly why we think that it is very well suited to investigate and to experiment with aspects of consciousness in artificial systems.
We will comment later on applying  Integrated Information Theory to the   automated reasoning system  Hyper in cases where it uses large knowledge bases. We will argue that those knowledge bases offer mechanisms for its integration and that this is necessary in order to do automated reasoning.

Refering to the small example with the camera pixels from the beginning of this section, it should be clear, that Integrated Information Theory is not only about handling large amounts of information, but also about integrating its parts.  
In the following  subsection we will depict another approach to consciousness, which can be seen as a special case of the Integrated Information Theory. It is  the  Global Workspace Theory developed by \textcite{Baars:97}, which is mainly focusing on the aspect of handling huge amounts of knowledge.

\subsection{The Theater of Consciousness}
\label{sec:GWT}
One motivation for Baars Global Workspace Theory \parencite{Baars:97} is the observation that the human brain has a very limited working memory. We can actively  manipulate  about seven separate things at the same time in our working memory. 
This is an astonishing small number in contrast to the more than 100 billion neurons of the human brain. Another limitation is that human consciousness is limited to only one single stream of input. We can listen only to one speaker at a time, we cannot talk to a passenger during driving in heavy traffic and there are many more examples like this. At the same time there are numerous processes running in parallel but unconsciously. Global Workspace Theory uses the metaphor of a theater to
 model how consciousness enables us  to handle the huge amount of knowledge, memories and sensory input the brain is controlling at every moment.

Global Workspace Theory assumes a theater consisting of a stage, an attentional spotlight shining at the stage, actors which represent the contents, an audience and some people behind the scene. Let's look at the parts in more detail:

\emph{The stage.} The working memory consists of verbal and imagined items. Most parts of the working
 memory are in the dark, but there are a few active items, usually the short-term memory. 

\emph{The spotlight of attention.} This bright spotlight helps in guiding and navigating through the working memory. Humans can shift it at will, by imagining things or events.

\emph{The actors} are the members of the working memory; they are competing against each other to gain access to the spotlight of attention.

\emph{Context behind the scene.}
Behind the scenes, the director coordinates the show and stage designers and make-up artists prepare the next scenes. 

\emph{The audience.} According to Baars, the audience represents the vast collection of specialized knowledge. It can be considered as a kind of long-term memory and consists of specialized properties, which are unconscious. Navigation through this part of the knowledge is done mostly unconsciously. This part of the theater is additionally responsible for interpretation of certain contents e.g., objects, faces or speech -- there are some unconscious automatisms running in the audience.

It is important to note that this model of consciousness, although it uses a  theater metaphor, is very different from a model like the Cartesian theater, as it it discussed and refused  in \textcite{Dennett1992-DENTAT}. A Cartesian theater model would assume that there is a certain region within the brain, which is the location of consciousness, this would be a kind of humunculus. In Baars theater, however, the entire brain is the theater and hence there is no special location, it is the entire integrated structure which is conscious. This is in very nice accordance with Tononi's Integrated Information Theory, described above.

Baars and colleagues also developed an architecture based on the Global Workspace Theory, which they call LIDA ``a comprehensive, conceptual and computational model covering a large portion of human cognition'' \parencite{Baars:LIDA}. 
LIDA consists of a number of software modules, which implement a cognitive cycle which is derived from the Global Workspace Theory. Baars gives a detailed justification of LIDA by modelling aspects of human cognition within his model. There are numerous other cognitive architectures which are used for different purposes, an extensive overview can be found in \textcite{kotseruba202040}.

In the following we will follow a very different road --- instead of designing a new architecture based on Global Workspace Theory, we will show that the reasoning system we have depicted and discussed in this paper
can be seen as an instance of Global Workspace Theory. It is important to note, that the development of this system was driven by the needs to be used as a reasoner within a commonsense system together with huge amounts of world knowledge. This system was not developed as a model for consciousness, but exhibits nevertheless strong similarity with the Global Workspace Theory and the Integrated Information Theory as we will demonstrate in the following subsection.

\subsection{Looking through the Global Workspace Theory-Glasses}
\label{sec:gwt-glasses}

We demonstrate how Hyper in combination with the associative selection within large knowledge bases can be interpreted as an architecture that implements the Global Workspace Theory as introduced in the previous section. 


\emph{The stage.} The working memory corresponds to the branch of the Hyper tree which is currently expanded. In the right part of Figure~\ref{fig:selection} this is the green path of the tree --- it contains the context in which the next reasoning step will be performed.

\emph{The spotlight of attention.} This bright spotlight selects and highlights those parts of the (green) branch together with the formulae from the problem or the selected parts of the  knowledge base which are used for the next  reasoning step.

\emph{The actors.} The actors correspond to the application of inference rules on the set of formulae currently processed by the theorem prover. The result of the actors' actions correspond to new formulae derived by an inference step. The spotlight of attention decides which actor, i.e. inference rule together with the necessary formulae, from the stage are to be active next.

\emph{Context behind the scene.}
Behind the scenes,  the reasoner and its control act as a director.

\emph{Audience.} According to Baars, the audience represents the vast collection of specialized knowledge. It can be considered as a kind of long-term memory, namely the   knowledge base. We depicted several selection mechanisms for finding the appropriate parts of the knowledge. 
The associative  selection mechanism uses word similarities from word embeddings. These word similarities connect different parts of the background knowledge with each other and thus allow to transfer existing knowledge to similar areas.
The selection mechanisms can be seen as the unconscious
interpretation skills which Baars located in the audience. In our case we only deal with declarative knowledge, if we had procedural knowledge as well, this would be part of the audience too.
\bigskip

Altogether we have a complete Baars' theater of consciousness consisting of the reasoner Hyper, together with its control and its background knowledge --- we have a system which can be interpreted as a  system with a certain degree of consciousness according to the ideas presented above. However, it should be noted that ``Metaphors must be used with care. They are always partly wrong, and we should keep an eye out for occasions when they break down. Yet they often provide the best starting point we can find.'' \parencite{Baars:97}

\section{Conclusion and Future Work} \footnotetext[7]{Figure~\ref{fig:snowman} was generated by automatically reducing the 300-dimensional GloVe word vectors to 2d vectors.} 

Levesque and other AI researchers consider the task to understand \emph{how} humans approach commonsense reasoning and \emph{how} this can be made available for AI-systems \emph{the real challenge for AI} \parencite{Levesque2017}, but it is even more a problem for cognitive modeling. Only if we understand the interplay between the cognitive processes or can -- at least show - how these processes can contribute to the respective competency, we understand better an important aspect of our (human) cognitive system. In our analysis, we have focussed on some parts of associative thinking and reasoning as well as on  commonsense reasoning and demonstrated that the presented approach can solve a variety of these tasks. For this we connected work from research on associative thinking with work on formal reasoning. We observed that AI systems in the commonsense reasoning domain face similar problems as humans in everyday reasoning: Both have to deal with large amounts of knowledge and have to be able to focus on the part of the knowledge which is relevant for the task at hand. To accomplish this task, humans rely on associative thinking. We argue that AI systems can use an instrument similar to associative thinking to deal with this problem as well.

 We depicted an implementation of a mind-wandering process  with associative reasoning. Further work has to be done for finding a way to determine what knowledge  is interesting enough to be kept within the focus of the system and how the knowledge base should be modified according to the results of mind-wandering. Currently, only one path of the proof tree is considered for the mind-wandering process. In future work we plan to extend the approach to consider multiple open branches for mind-wandering.

We have given a brief insight into how associative reasoning with large amounts of knowledge also play an important role in the field of creativity. In future work, we will not only consider selection of background knowledge in the area of creativity, but also investigate how automated reasoning can be used in this area.

Finally we broadend our view by discussing the  associative reasoning system as a conscious system in the sense of the Integrated Information Theory and the Global Workspace Theory. Future work on this topic should relate all these observations to discussions from the area of theory of mind --- one of the great cognitive modeling challenges.

\printbibliography


\end{document}